\documentclass[conference, letterpaper, final]{IEEEtran}

\usepackage{cite}
\usepackage{amsmath,amssymb,amsfonts}
\usepackage{algorithmic}
\usepackage{listings}
\usepackage{graphicx}
\usepackage{textcomp}
\usepackage{xcolor}
\usepackage{hyperref}
\usepackage{svg}
\usepackage{placeins}
\usepackage[margin=0.8in]{geometry}
\usepackage{caption}
\usepackage{subcaption}
\usepackage{graphicx}
\usepackage{xspace}
\usepackage{acronym}
\usepackage{tabularx} 
\usepackage{array}
\usepackage{ragged2e} 
\usepackage{todonotes}
\usepackage{comment}

\DeclareRobustCommand*{\IEEEauthorrefmark}[1]{%
  \raisebox{0pt}[0pt][0pt]{\textsuperscript{\footnotesize #1}}%
}

\def\BibTeX{{\rm B\kern-.05em{\sc i\kern-.025em b}\kern-.08em
    T\kern-.1667em\lower.7ex\hbox{E}\kern-.125emX}}

\definecolor{mygreen}{rgb}{0,0.4,0}
\definecolor{mygray}{rgb}{0.5,0.5,0.5}
\definecolor{mymauve}{rgb}{0.2941,0,0.5098}
\definecolor{flame}{rgb}{0.89, 0.35, 0.13}

\newcommand{\chiara}[1]{\textcolor{black}{#1}} 

\newcommand{\ourmethod}{AcME-AD\xspace}
\acrodef{PIADE}[PIADE]{Packaging Industry Anomaly DEtection}
\acrodef{AD}[AD]{Anomaly detection}
\acrodef{IF}[IF]{Isolation Forest}
\acrodef{TEP}[TEP]{Tennessee Eastman Process}
\acrodef{LODA}[LODA]{Lightweight on-line detector of anomalies}

\DeclareFixedFont{\ttb}{T1}{txtt}{bx}{n}{9} 
\DeclareFixedFont{\ttm}{T1}{txtt}{m}{n}{9}  
\begin{document}

\title{\vspace{0.25in} Enabling Efficient and Flexible Interpretability of Data-driven Anomaly Detection in Industrial Processes with AcME-AD}

\author{\IEEEauthorblockN{
    Valentina Zaccaria\IEEEauthorrefmark{1},
    Chiara Masiero\IEEEauthorrefmark{2},
    David Dandolo\IEEEauthorrefmark{2},
    Gian Antonio Susto\IEEEauthorrefmark{1}
    }

\IEEEauthorblockA{
    \IEEEauthorrefmark{1}Department of Information Engineering, University of Padova,\,
    \IEEEauthorrefmark{2}Statwolf Data Science Srl
}
\thanks{Corresponding author: valentina.zaccaria@unipd.it}
\thanks{This work was partially carried out within the MICS (Made in Italy – Circular and Sustainable) Extended Partnership and received funding from Next-GenerationEU (Italian PNRR – M4C2, Invest 1.3 – D.D. 1551.11-10-2022, PE00000004). Moreover this study was also partially carried out within the PNRR research activities of the consortium iNEST (Interconnected North-Est Innovation Ecosystem) funded by the European Union Next-GenerationEU (Piano Nazionale di Ripresa e Resilienza (PNRR) – Missione 4 Componente 2, Investimento 1.5 – D.D. 1058 23/06/2022, ECS00000043).}
}

\maketitle
\begin{abstract}
While Machine Learning has become crucial for Industry 4.0, its opaque nature hinders trust and impedes the transformation of valuable insights into actionable decision, a challenge exacerbated in the evolving Industry 5.0 with its human-centric focus. This paper addresses this need by testing the applicability of \ourmethod in industrial settings. This recently developed framework facilitates fast and user-friendly explanations for anomaly detection. \ourmethod is model-agnostic, offering flexibility, and prioritizes real-time efficiency. Thus, it seems suitable for seamless integration with industrial Decision Support Systems. We present the first industrial application of \ourmethod, showcasing its effectiveness through experiments. These tests demonstrate \ourmethod's potential as a valuable tool for explainable AD and feature-based root cause analysis within industrial environments, paving the way for trustworthy and actionable insights in the age of Industry 5.0.
\end{abstract}


\section{Introduction}

The convergence of sensing, actuation, and communication technologies, coupled with the rise of Machine Learning (ML), has spurred the rapid development of the Internet of Things (IoT) \cite{bargellesi2021autoss, saleem_2021, KANG2020106773}. Within this landscape, Industrial IoT (IIoT) stands out, where continuous monitoring of industrial assets is crucial for optimizing maintenance schedules and maximizing production efficiency.

\ac{AD} plays a vital role in IIoT systems, enabling the timely identification of deviations from normal behavior before they escalate into costly downtime or production issues \cite{brito2022explainable}. Notably, unsupervised AD techniques are particularly advantageous in this domain, as they do not necessitate pre-labeled data, often a cumbersome and resource-intensive endeavor.

However, merely identifying anomalies is insufficient. Understanding the underlying causes behind them is critical for implementing effective corrective actions. This is where explainable Artificial Intelligence (XAI) comes into play \cite{expl, adadi_2018, Carvalho_2019}. XAI, also known as ML interpretability, aims to elucidate the decision-making process of ML models. While some ML methods offer inherent interpretability, others, like high-performance deep neural networks, often operate as "black boxes," lacking transparency in their predictions. This lack of transparency can lead to hesitation in adopting ML solutions within Decision Support Systems (DSS) due to potential trust concerns \cite{Nicodeme2020}.

In the context of IIoT, fast and efficient XAI is critical. Rapid explanation generation is crucial for effective decision-making and prompt implementation of corrective actions, minimizing potential delays in response.

This paper examines how \ourmethod \cite{acme_ad} an efficient, and model-agnostic approach to explainability that was recently developed, can be applied in industrial settings for AD.

\ourmethod works well even with unsupervised AD, as it explains predictions by highlighting how changes in input features impact the resulting anomaly score. This, combined with its computational efficiency and effective result visualizations, makes \ourmethod ideal for industrial internet of things (IIoT) monitoring through decision support systems (DSS).

The rest of this paper is organized as follows:
Section \ref{sec:related_works} explores the critical role of XAI in Industry 4.0 and 5.0, particularly within AD.
Section \ref{sec:proposed_approach} summarizes the core principles and functionalities of \ourmethod.
Section \ref{sec:experimental_results} showcases the effectiveness of \ourmethod through real-world applications in two industrial scenarios.
Section \ref{sec:conclusions} recaps key takeaways and outlines promising directions for future exploration.

\section{Related Work}\label{sec:related_works}
\chiara{
Recent advancements in data and computational capabilities enabled the widespread adoption of ML within digital manufacturing. Transitioning towards Industry 5.0, the focus shifts to a human-centric paradigm, emphasizing collaboration between operators and intelligent systems \cite{Ahmed2022FromAI}.  Operators evolve from primarily executing tasks to leveraging their domain knowledge alongside AI-powered tools for informed decision-making. Consequently, the need for explainable and trustworthy predictions becomes paramount for building confidence in data-driven monitoring solutions, making ML interpretability a critical factor.
\\
Examples of explainable ML industrial solutions include fault detection in machinery \cite{brito2022explainable}, process monitoring in various industries from semiconductor  \cite{carletti2020interpretable, feng_2020} to home appliances \cite{carletti2019deep}, and predictive maintenance \cite{simon_2021}.
\\
This work focuses on interpretable anomaly detection. Prior research on this topic spans diverse areas like wireless spectrum anomalies  \cite{raje_2019}, robot activity recognition \cite{hayes_2017}, and ICT threat detection \cite{orizio_2020}.
\\
To balance low latency with root-cause analysis insights, we leverage \ourmethod \cite{acme_ad} to explain predictions. This approach maintains efficiency and model-agnosticism in line with AcME \cite{dandolo2023acme}. However, \ourmethod is uniquely tailored for (possibly unsupervised) anomaly detection (AD) tasks, assuming only that the model provides an anomaly score as output. This combination of explainability and efficiency positions \ourmethod as an ideal solution for industrial scenarios where timely and actionable insights are essential.
}

\section{Proposed Approach}\label{sec:proposed_approach}
\chiara{
The goal of this work is to prove the effectiveness of \ourmethod \cite{acme_ad} 
to explain AD models in the industrial scenario.   
\ourmethod offers a model-agnostic approach to explainability in AD, contrasting methods tailored for specific models like Isolation Forest \cite{carletti2023interpretable} or for Principal Component Analysis in \cite{takeishi2019shapley}. This flexibility allows users to leverage the best-performing AD model while gaining interpretable results from \ourmethod.
\\
Furthermore, \ourmethod prioritizes computational efficiency, making it suitable for critical scenarios requiring real-time decisions. This distinguishes it from methods like SHAP \cite{lundberg2017unified} which can struggle in such settings. As shown in Section \ref{sec:experiments}, \ourmethod delivers explanations quickly, solidifying its value for time-sensitive applications.
\\
}
\chiara{
\ourmethod produces local explanations for a data point $\mathbf{x}$ by systematically perturbing individual features, while keeping other features fixed, and observing the impact on the anomaly score. It utilizes a convex combination of four key metrics  (all with range from 0 to 1) to compute $I_j(\mathbf{x)}$, the importance of the $j-$feature in contributing to the anomaly score assigned to $\mathbf{x}$:
\begin{enumerate}
\item Delta $D_j$: Represents the maximum difference in anomaly scores achieved by perturbing feature $j$ across its quantile values. 
\item Ratio $R_j$: Represents the normalized distance of the original anomaly score from the minimum achievable score through $j$'s perturbation. 
\item Change of Predicted Class $C_j$: $C_j=1$ if feature perturbation induces a change in the classification (anomaly vs normal) of the data point $\mathbf{x}$.
\item Distance to Change $Q_j$:  Quantifies the degree of perturbation needed (in terms of quantile value) for feature j to change the classification outcome of $\mathbf{x}$, if possible. 
\end{enumerate}
Then
    \begin{gather*}
          I_j(\mathbf{x})  \doteq w_D D_j + w_C C_j + w_Q Q_j + w_R R_j \\
        \text{s.t.} \; \sum_{i} w_i = 1, \; w_i \geq 0,  \; \forall i \in \{D, C, Q, R\}.
    \end{gather*}
The weights $w_i$ are input parameters and can be customized. Default values are set to $w_D = 0.3, w_C = 0.3, w_R =0.2, w_Q = 0.2$.
\\
\ourmethod's 'what-if` visualization tool (see for example Fig. \ref{fig:loc_tep}) provides local explanation by exploring how the anomaly score would change if a single feature's quantile value were modified while keeping others fixed. The ten most important features, ranked by decreasing importance, are displayed. The visualization features: (i) a red dashed line representing the analyzed data point's anomaly score; (ii) a black solid line indicating the normal/anomalous classification threshold, and (iii) large colored bubbles on the red line depicting the data point's actual feature values. These bubble colors convey the position of these values within the entire dataset's distribution (specifically, their quantile).
By interacting with this visualization, users can understand how modifying specific feature values might impact the anomaly score and potentially transition the data point from anomalous to normal.
\\
To identify overall important features (i.e., relevant across different anomalies), we propose the following procedure: (i) Isolate anomalies: Select all data points flagged as anomalies by the model; (ii) Localize importance: For each anomaly, employ AcME-AD to generate a local feature ranking based on importance scores; (iii) Aggregate ranks: Count how often each feature appears at each rank position across all anomalies; (iv) Visualize insights: Normalize the counts and create a stacked bar chart (see, for example Fig. \ref{fig:tep_global_if}).
\\
The x-axis represents ranking positions (1 being most important). Each bar at position k for feature j reflects the percentage of anomalies where feature j ranked kth. This allows researchers to observe which features consistently rank high across various anomalies.
For many features, consider merging those ranking below 5\% into an "others" category to improve readability. This threshold can be adjusted based on specific needs.}

\section{Experimental Results}\label{sec:experimental_results}
\label{sec:experiments}
This section presents the experimental results obtained using \ourmethod in two industrial scenarios. Experiments involve both a synthetic dataset and an unlabeled real-world dataset. The former provides known anomaly-causing features for initial validation, while the latter reflects real-world challenges such as high feature dimensionality and lack of supervision.
The code to reproduce the experiments is available in a public repository\footnote{\url{https://github.com/dandolodavid/ACME}}.

\subsection{Use Case I: Chemical processes}
In the first case-study, we investigate the effectiveness of \ourmethod using the \ac{TEP} dataset\cite{rieth2017additional}, a well-established industrial benchmark for fault and anomaly detection. This dataset, being supervised, enables us to select a high-performing \acs{AD} model to explain, crucial to properly assessing our method.  
\chiara{
Since \ourmethod explains model predictions and not the underlying physical process, it is important to pick a very accurate model to leverage domain knowledge about the process in evaluating interpretability. In principle, indeed, misattributions of feature importance could be associated with shortcomings in the model rather than limitations in the interpretability approach.}

\ac{TEP} data are obtained by simulation using a computational model of chemical processes. 
The dataset comprises time series originating from normal processes or faulty processes. Each time series contains 500 samples with 52 features. Samples are categorized into 21 classes, where Class 0 signifies normal operational states and classes 1-20 correspond to different simulated process faults. Importantly, there exists prior knowledge about the features associated with each of the first 15 faults \cite{don2019dynamic}, aiding the assessment of the proposed method.

In our experimental setup, we consider a subset of the available dataset to reproduce an AD scenario. We randomly select 70 normal simulations and 3 faulty simulations of fault type \texttt{IDV12}. We select this fault type because domain knowledge about the responsible feature is available \cite{don2019dynamic,harinarayan2022xfddc}. Specifically, the fault \texttt{IDV12} is related to the condenser cooling water input temperature, which assumes random anomalous values during the process. The true root cause feature is \texttt{xmeas\_11}, which is the separator temperature measure. Among the trained models, \ac{IF} \cite{liu2008isolation} is the best performing one, achieving an average precision score equal to 0.77. Results for other models can be found in the code repository. 

\subsubsection{Overall feature importance}
\begin{figure}
    \centering
    \includegraphics[width=1\linewidth]{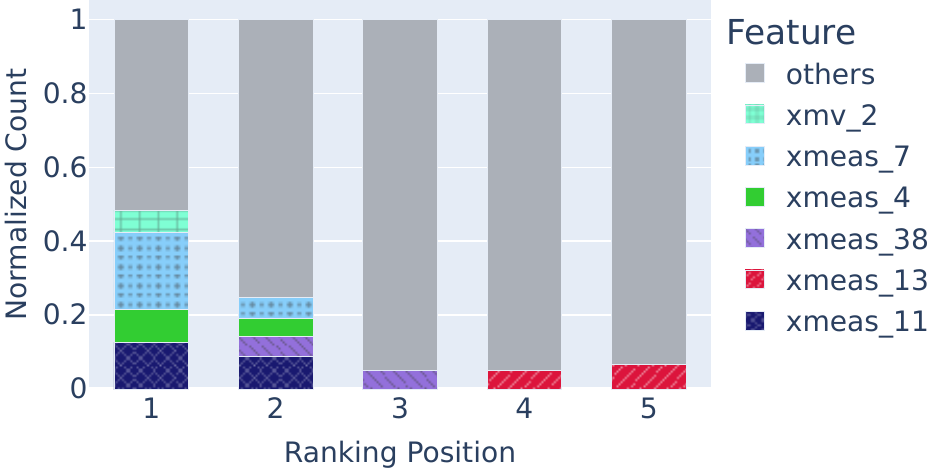}
    \caption{\ourmethod overall importance bar chart using \ac{IF} on \ac{TEP}.}
    \label{fig:tep_global_if}
\end{figure}

\begin{figure}
    \centering
    \includegraphics[width=1\linewidth]{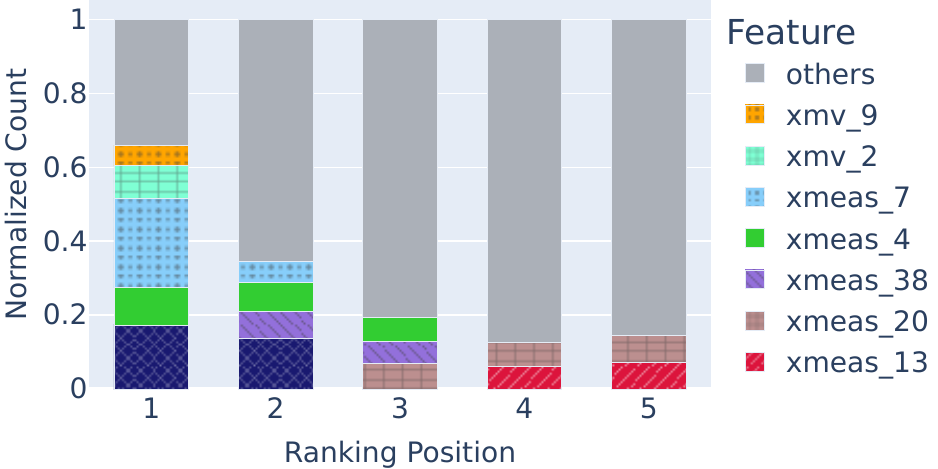}
    \caption{KernelSHAP overall importance bar chart using \ac{IF} on \ac{TEP} and 10\% sampling of the background dataset.}
    \label{fig:tep_global_kernelshap}
\end{figure}

Fig. \ref{fig:tep_global_if} displays the overall importance bar chart for \ac{TEP}. Feature \texttt{xmeas\_11}, depicted in dark blue, is correctly identified as the most or second most relevant feature in a considerable percentage of anomalies, consistest with prior knowledge. 
It is second in importance only to \texttt{xmeas\_7}, which is the reactor pressure. The authors of \cite{don2019dynamic}, in Section 4, builded a Sign Directed Graph to model the \acl{TEP}. This graph shows that reactor pressure \texttt{xmeas\_7} influences \texttt{xmeas\_11} through only another feature, suggesting that \ac{IF} might leverage this causal relationship to detect anomalies. 
\\
\textcolor{black}{Consistent with prior research \cite{dandolo2023acme, acme_ad}, KernelSHAP \cite{lundberg2017unified} and AcME explanations show high similarity. This is evident when comparing Figures~\ref{fig:tep_global_if} and~\ref{fig:tep_global_kernelshap}. However, computing a local explanation takes $3.03$ s with \ourmethod and $62.32$ s with KernelSHAP (as detailed in Tab. \ref{tab:time_comparison_TEP}). 
}

\subsubsection{Local Interpretability}
\begin{figure*}
    \centering
    \includegraphics[width=\linewidth]{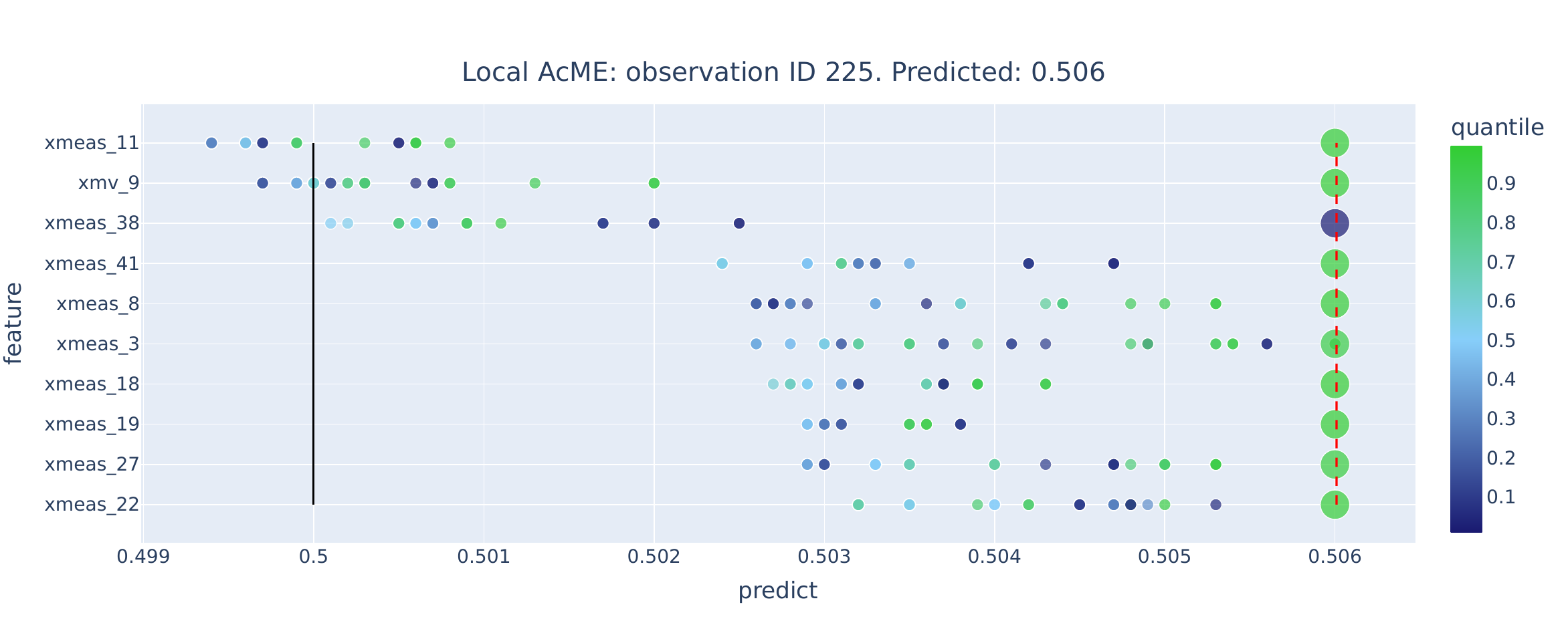}
    \caption{\ourmethod Local explanation of an anomaly identified by \ac{IF} in \ac{TEP}. It emerges that this data point would be considered normal by modifying the values of \texttt{xmeas\_11} or \texttt{xmv\_9}. 
    }
    \label{fig:loc_tep}
\end{figure*}

Fig. \ref{fig:loc_tep} illustrates \ourmethod \textit{what-if} tool for local explanation. For this specific anomalous prediction, the feature \texttt{xmeas\_11}, representing the separator temperature measurement, correctly emerges as the most relevant. Its actual value, denoted by the larger green bubble, is very high. However, one can notice that reducing the separator temperature would transition the point to a normal state, at least up to a certain quantile value, represented by the leftmost light blue point. After that, further decreases in the separator temperature would classify the data point as abnormal again.  
The only other feature that, if perturbed, would induce a change of state is \texttt{xmv\_9}, representing the stripper steam valve, a manipulated process variable. The pattern exhibited by this feature closely resembles that of \texttt{xmeas\_11}.

\subsection{Use Case II: Packaging equipment}
The second industrial case-study considers the monitoring of packaging machines. The experiments are conducted on the publicly available\footnote{https://zenodo.org/records/7071747} \ac{PIADE} dataset \cite{diego2022packaging}, which contains unlabeled data acquired by five industrial packaging machines. 
We consider sequences data, where raw observations are aggregated considering 1-hour-long time windows. Because the the five machines have different operating points, in this paper, we present results obtained considering the second machine with \texttt{Equipment\_ID} equal to 2. Results for other machines can be found in the repository. The dataset contains 2725 data points, each one characterized by 162 features.

\subsubsection{Model selection with \ourmethod}
\label{sec:modelselection}
In IIoT datasets, such as in \ac{PIADE}, labels are often not available. 
Typically, practitioners resort to ensembles of \ac{AD} models. When one of the models detects an anomalous data point, an alarm is triggered. However, this approach often generates a considerable number of false positives. 
\chiara{Instead, \ourmethod can be exploited to select a suitable AD model.
This selection process is informed by the features identified as relevant for each model according to AcME-AD.} This opportunity is strictly related to the model-agnostic nature of \ourmethod.

\begin{figure}
    \centering
        \includegraphics[width=\linewidth]{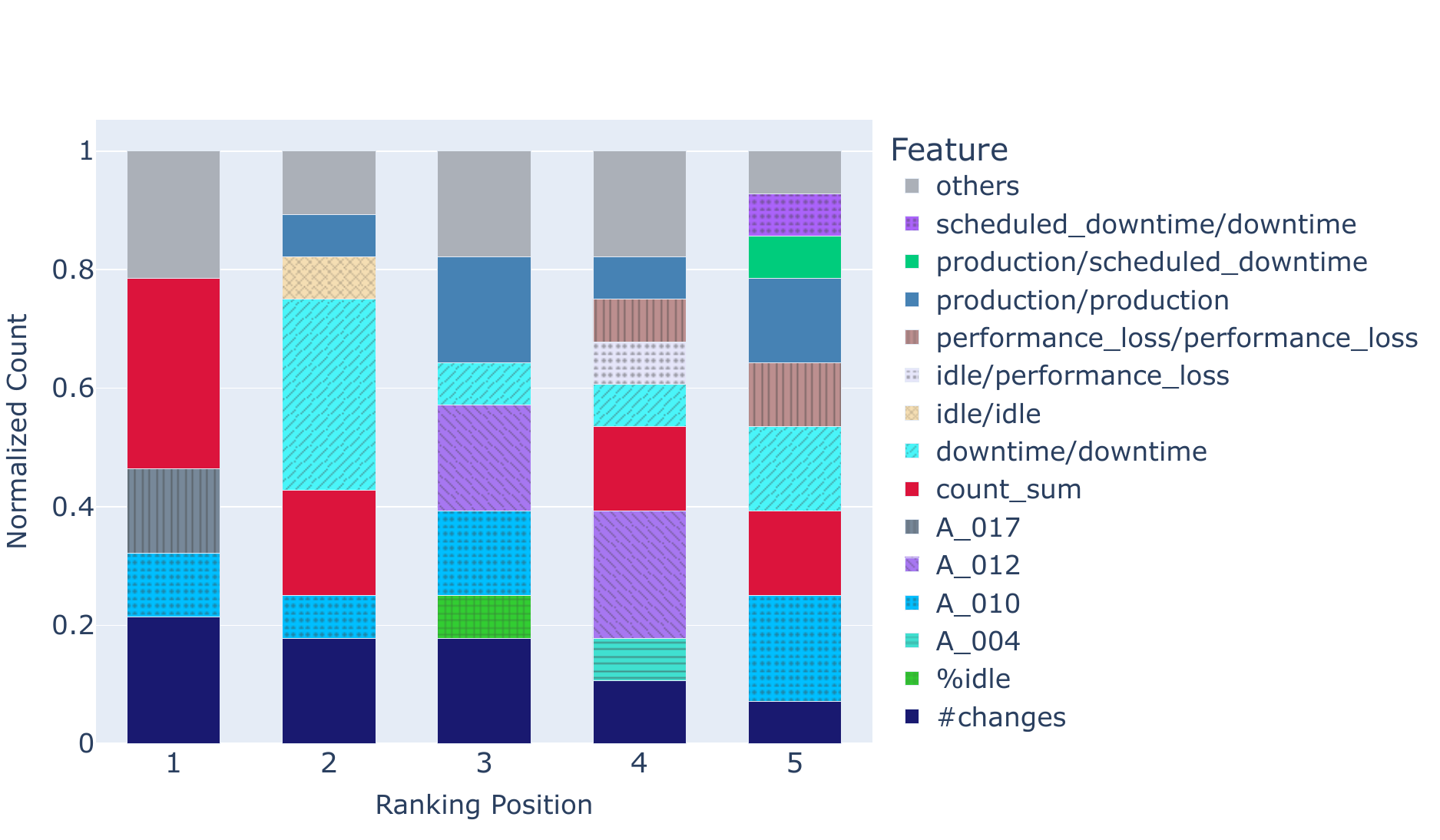}
    \caption{Overall importance bar chart for LODA on \ac{PIADE} sequences - Equipment 2.}
    \label{fig:piade_model_comparison}
\end{figure}

\chiara{We compute the overall importance for a set of candidate AD models as detailed in Section \ref{sec:proposed_approach}. Then, in collaboration with subject matter experts, it is possible to select \ac{LODA} \cite{pevny2016loda} as the model that best aligns with their domain knowledge about the underlying process.} Fig. \ref{fig:piade_model_comparison} shows the resulting overall importance visualization. As expected by the domain experts, the most relevant features are the count of alarms \texttt{count\_sum} and the number of state changes \texttt{\#changes}, followed by specific alarms such as \texttt{A\_017} and \texttt{A\_010} that are related to known failures, and \texttt{downtime/downtime} that is related to persistent downtime status.

\begin{figure*}[t]
    \centering
    \includegraphics[width=1\linewidth]{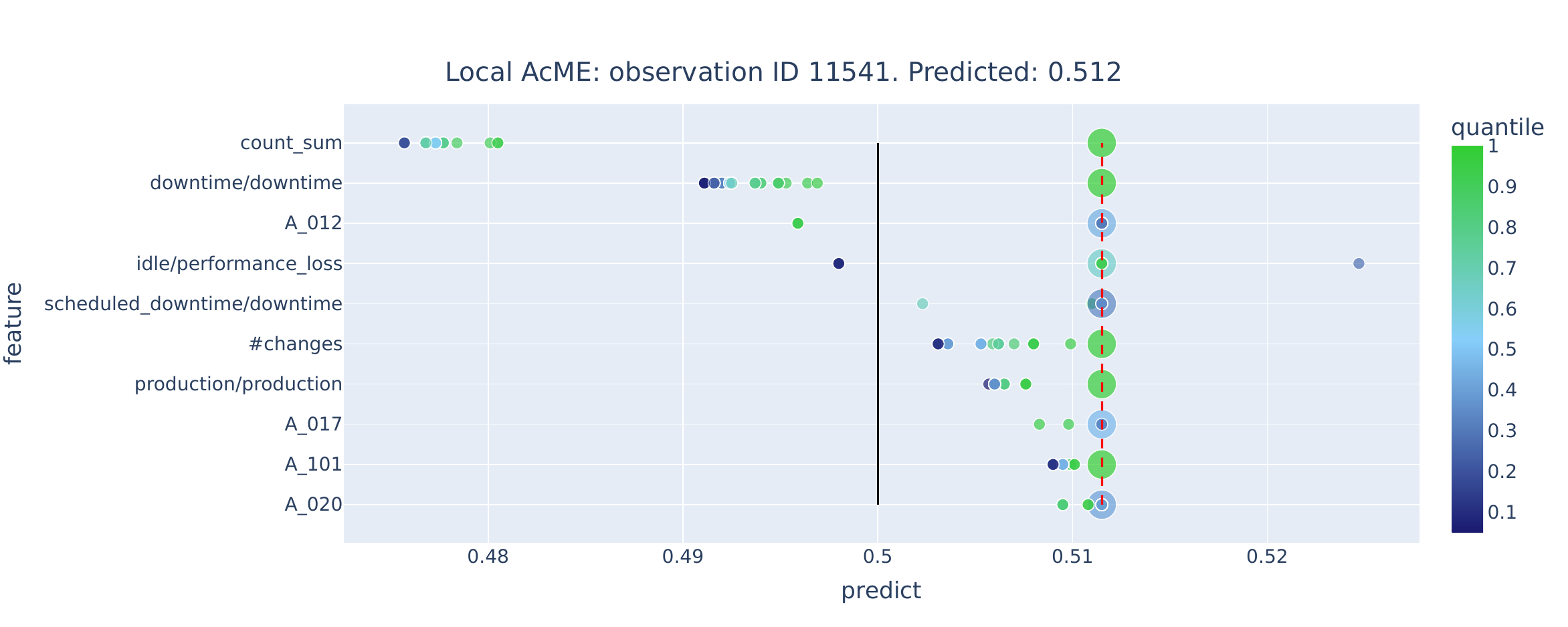}
    \caption{\ourmethod Local explanation of a randomly sampled anomaly identified by \ac{LODA} in \ac{PIADE}.}
    \label{fig:loc_piade_equip2_loda}
\end{figure*}

\subsubsection{Local Interpretability}
As a demonstrative example, let's consider a randomly sampled data point detected as anomalous by \ac{LODA}. We generate the corresponding \ourmethod's \textit{what-if} visualization, displayed in Fig. \ref{fig:loc_piade_equip2_loda}. 
There are four features that, if perturbed, would bring the data point to a normal state. Among these, the most important is \texttt{count\_sum}. We recall that this feature counts the number of alarms occurred within the 1-hour interval described by the data point. Its original value, represented by the larger green bubble, falls within the highest quantile of the training set, being anomalous on its own. 
The other smaller bubbles show the anomaly scores resulting from perturbations of \texttt{count\_sum} across various quantiles, while keeping the other features fixed. As the \texttt{count\_sum} value decreases, depicted by the gradual change in color of the bubbles from green to blue, the data point tends toward normality. This behavior aligns with domain knowledge, where a high number of alarms typically indicates issues in the machine, thus being anomalous situations. Notably, even a slight perturbation of \texttt{count\_sum} leads to the classification of the data point as normal. 

The second most relevant feature is \texttt{downtime/} \texttt{downtime}. The anomaly score behavior closely mirrors that of \texttt{count\_sum}. This 1-hour-sequence data point exhibits a high frequency of transitions from downtime to downtime, which is indeed atypical. 

The third feature which perturbation can induce a classification change is \texttt{A\_012}. Many occurrences of this process alarm are associated with high production levels in the equipment under consideration. Consequently, frequent occurrences of this alarm can be interpreted as an indication of a normal and desirable operational situation. 
The current value of \texttt{A\_012} has a light blue color, suggesting a significantly reduced machine throughput compared to the majority of all the data points. The \textit{what-if} tool points out that if the number of \texttt{A\_012} increases (the green bubble), the data point would return to a normal classification, which is aligned with domain expertise.

The last feature that can be perturbed to normalize the sequence is \texttt{idle/performance\_loss}, which counts the number of machine transitions from idle to a performance loss state within the interval. However, the absence of a correlation between the values assumed by this feature and the corresponding anomaly score suggests that the feature can not be effectively utilized for taking corrective actions. 

\chiara{This analysis demonstrates how AcME-AD empowers users to pinpoint feature-specific interventions that can potentially bring anomalous data points back to normal, aiding in informed decision-making and system restoration.}

\subsection{\ourmethod and KernelSHAP: time comparison}
\label{subsec:timecomparison}
The \textit{de-facto} standard model-agnostic method for explaining models, including unsupervised \ac{AD} ones, is KernelSHAP \cite{lundberg2017unified}. Despite its solid theoretical foundation, the computational overhead of KernelSHAP is a recognized issue \cite{dandolo2023acme}. Specifically, the computational time quickly escalates with the size of the dataset used to fit the explainer, denoted as \textit{background}, which is typically equal to the training set.
The authors of \cite{APISHAP} propose to reduce the size of the background by randomly sub-sampling the original dataset. However, as extensively demonstrated in \cite{dandolo2023acme}, this operation may lead to unreliable explanations, especially in complex dataset distributions. 
Table \ref{tab:time_comparison_TEP} reports the time\footnote{For reference, experiments are conducted on AMD Ryzen 7 3700X, 3.6GHz, RAM 20GB.} required to compute a local explanation of a single anomalous data point using \ourmethod and KernelSHAP on \ac{TEP}. Due to limited resources, we were able to compute only the time of KernelSHAP using at most half of the training set as background. Our findings reveal that \ourmethod is an order of magnitude faster than KernelSHAP even when using the 5\% as background. When utilizing the 20\% and the 50\% sampling, the time demand makes KernelSHAP incompatible with the integration in Decision Support Systems.

Table \ref{tab:time_comparison_PIADE} illustrates the elapsed time when analyzing \ac{PIADE} data. In this case, the gap is less pronounced. The motivation behind this lies in the fact that, for the same number of features, the time complexity of KernelSHAP is dominated by the number of data points, whereas the complexity of \ourmethod is driven by the dimensionality of the samples (i.e., the number of features). \ac{PIADE} has a high number of features (162) and a low number of samples (2725) when compared to \ac{TEP} (36500). Nevertheless, once again, \ourmethod generally outperforms KernelSHAP in terms of speed, achieving comparable performance only for 25\% sampling. 
Moreover, this discrepancy becomes substantial when exploiting the explainability method for model selection, as in Section \ref{sec:modelselection}, where local explanations need to be generated for every predicted anomalous point.

Finally, it is noteworthy that there exist model-specific variants of SHAP which, by leveraging the specific structure of the models, demand lower computational time. However, in this comparison, we solely focus on KernelSHAP to ensure a fair evaluation of \ourmethod, which is model-agnostic. 

\begin{table}[t]
    \centering
    \setlength{\extrarowheight}{5pt}
    
    \begin{tabular}{p{0.2\linewidth}>{\raggedleft\arraybackslash}p{0.30\linewidth}>{\raggedleft\arraybackslash}p{0.35\linewidth}}

         & \textbf{Background Size} &\textbf{Elapsed time IF (s)}\\
        \hline
        \ourmethod      &   36500 (100\%)    &   3.03  \\
        KernelSHAP      &   1825 (5\%)      &   30.23   \\
        KernelSHAP      &   3650 (10\%)     &   62.32     \\                             
        KernelSHAP      &   7300 (20\%)     &   123.75     \\
        KernelSHAP      &   36500 (50\%)    &    655.64\\
        \hline
    \end{tabular}
    \caption{Single explanation times for \ourmethod and KernelSHAP with \ac{TEP} dataset.}
    \label{tab:time_comparison_TEP}
\end{table}

\begin{table}[t]
    \centering
    \setlength{\extrarowheight}{5pt}
    
    \begin{tabular}{p{0.2\linewidth}>{\raggedleft\arraybackslash}p{0.30\linewidth}>{\raggedleft\arraybackslash}p{0.35\linewidth}}

         & \textbf{Background Size} &\textbf{Elapsed time LODA (s)}\\
        \hline
        \ourmethod      &   2725 (100\%)    &   18.63   \\
        KernelSHAP      &   681 (25\%)      &   16.46   \\
        KernelSHAP      &   1362 (50\%)     &   33.21   \\                             
        KernelSHAP      &   2043 (75\%)     &   49.58   \\
        KernelSHAP      &   2725 (100\%)    &   65.64   \\
        \hline
    \end{tabular}
    \caption{Single explanation times for \ourmethod and KernelSHAP with \ac{PIADE} dataset.}
    \label{tab:time_comparison_PIADE}
\end{table}

    


\section{Conclusions}\label{sec:conclusions}
This work successfully demonstrates the application of our proposed method, \ourmethod, for anomaly detection in an industrial setting.
\\
Firstly, \ourmethod facilitates user decision-making through clear and informative visualizations. The integrated "what-if" tool for local interpretation empowers users to pinpoint the root cause of anomalies and take appropriate corrective actions. Secondly, its efficient computation enables seamless integration into decision support systems, promoting real-time anomaly detection and response.
\\
We highlight the particular suitability of \ourmethod for unsupervised anomaly detection tasks, which are of paramount importance in industrial applications. By providing model-agnostic explanations, \ourmethod empowers domain experts to select models that best align with their knowledge, fostering crucial human-machine collaboration in the context of Industry 5.0. As future development of the present work, we aim at refining the visualizations based on feedback from real-world deployments. Additionally, we plan to explore the potential of utilizing local explanations for clustering anomalies into distinct groups, thereby enabling supervised classification tasks.

\section*{Acknowledgments}
This work was partially carried out within the MICS (Made in Italy – Circular and Sustainable) Extended Partnership and received funding from Next-GenerationEU (Italian PNRR – M4C2, Invest 1.3 – D.D. 1551.11-10-2022, PE00000004). Moreover this study was also partially carried out within the PNRR research activities of the consortium iNEST (Interconnected North-Est Innovation Ecosystem) funded by the European Union Next-GenerationEU (Piano Nazionale di Ripresa e Resilienza (PNRR) – Missione 4 Componente 2, Investimento 1.5 – D.D. 1058 23/06/2022, ECS00000043). This work was also cofunded by the European Union in the context of the Horizon Europe project ‘AIMS5.0 - Artificial Intelligence in Manufacturing leading to Sustainability and Industry5.0’ Grant agreement ID: 101112089. The Regione Veneto is also gratefully acknowledged for co-financing part of this activity through the Development and Cohesion Plan (PSC) under the special section THEME AREA PSC 1 - RESEARCH AND INNOVATION, in continuity with ACTION (POR FESR) 1.1.4 'Support for collaborative R\&D activities for the development of new sustainable technologies, products, and services,' as per DGR no. 1800 dated December 15, 2021 - Project MIMIC.

\bibliographystyle{IEEEtran}
\bibliography{citations}{}

\end{document}